\documentclass[conference]{IEEEtran}
\IEEEoverridecommandlockouts
\usepackage[numbers,sort&compress]{natbib}
\usepackage{amsmath,amssymb,amsfonts}
\usepackage{algorithmic}
\usepackage{graphicx}
\usepackage{textcomp}
\usepackage{xcolor}
\usepackage{url}
\usepackage{booktabs} % For professional lines
\usepackage{float}    % For the [H] placement specifier
\usepackage{array}
\usepackage{hyperref}
\hypersetup{
    colorlinks   = true,
    citecolor    = blue,
    urlcolor    =  blue
}
\newcommand{\dataSetName}{\texttt{SalAngaBhava}}
\newcommand{\footURL}[1]{\footnote{\url{#1}}}
\usepackage{makecell}
\usepackage{hhline}
\usepackage{tabularx}
    \newcolumntype{L}{>{\raggedright\arraybackslash}X}
    \newcolumntype{C}{>{\centering\arraybackslash}X}
    \newcolumntype{R}{>{\raggedleft\arraybackslash}X}
\usepackage{multirow}

\newcommand{\hf}[2]{\raisebox{-2.2pt}{\includegraphics[scale=0.09]{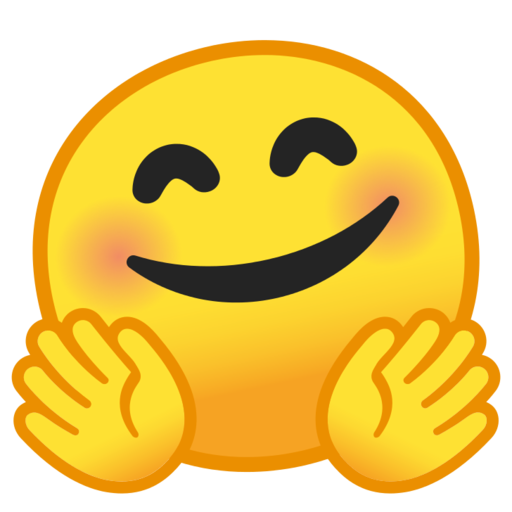}}~\href{#1}{\texttt{#2}}}

\newcommand{\gh}[2]{\raisebox{-2.2pt}{\includegraphics[scale=0.02]{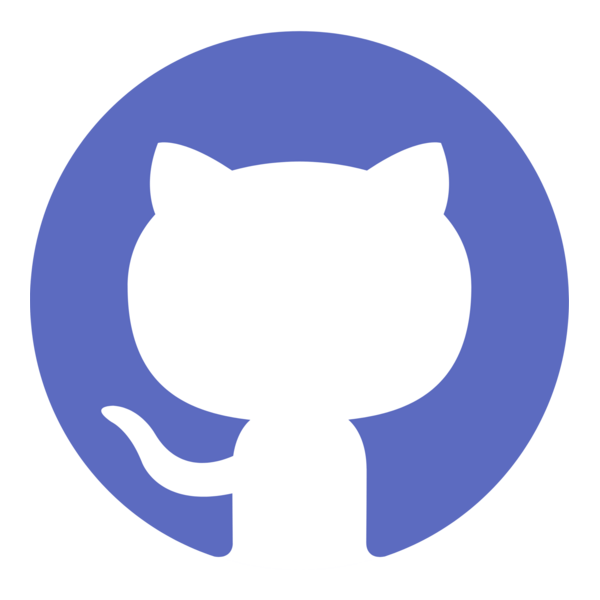}}~\href{#1}{\texttt{#2}}}
    
\def\BibTeX{{\rm B\kern-.05em{\sc i\kern-.025em b}\kern-.08em
    T\kern-.1667em\lower.7ex\hbox{E}\kern-.125emX}}
\begin{document}

\title{\dataSetName: A Sinhala Market Dataset for Aspect-based Sentiment Analysis}

\author{
\IEEEauthorblockN{%
Lakshani Galwatta\IEEEauthorrefmark{1},
Nisansa de Silva\IEEEauthorrefmark{1},
Sarangi Aththanayake\IEEEauthorrefmark{1},
Adithya Galwatta\IEEEauthorrefmark{2}}
\IEEEauthorblockA{\IEEEauthorrefmark{1}Dept.\ of Computer Science \& Engineering, University of Moratuwa, Sri Lanka.\\
\texttt{\{lakshani.25, NisansaDds, sarangi\}@cse.mrt.ac.lk}}
\IEEEauthorblockA{\IEEEauthorrefmark{2}School\ of Technologies, Cardiff Metropolitan university, United Kingdom.\\
\texttt{st20315307@outlook.cardiffmet.ac.uk}
}
}  

\maketitle

\begin{abstract}
Sentiment analysis has been a primary domain under Natural Language Processing (NLP) from its inception as it plays a vital role in both real-world and research applications. In high-resource languages, this has been extended a step further, and instead of predicting sentiment at the sentence level, models have been developed to detect more fine-grained sentiments at aspect level. However, in order to conduct this fine-grained Aspect-based Sentiment Analysis (ABSA), datasets annotated with aspects and sentiments toward the said aspects is required. Such datasets are lacking for low-resources languages among which, we can count Sinhala, an Indo-Aryan languages used primarily in Sri Lanka.  
%Fine-grained aspect-based sentiment analysis (ABSA) datasets are necessary to identify sentiment toward aspects within a text. But for low-resource languages such as Sinhala, such annotated datasets are scarce, which slows down research in fine-grained sentiment analysis in Sinhala Language. 
In this work, we introduce, \dataSetName, a new Sinhala Aspect-based Sentiment Analysis dataset which contains Sinhala product reviews that are manually labeled with aspect terms and the associated sentiments (positive, negative, neutral). 
% In this work, we introduce a new Sinhala Aspect-based Sentiment Analysis dataset to facilitate low-resource research in sentiment analysis. The dataset contains Sinhala reviews that are manually labeled with aspect terms and the associated sentiments (positive, negative, neutral). 
The data was collected from domain-relevant sources such as user-generated reviews and comments, and was annotated following carefully defined guidelines to ensure consistency and quality. The dataset consists of sentences and aspect–sentiment pairs, encompassing a considerable range of aspects from several domains. The analysis confirms that the dataset is well-structured and sufficiently balanced for ABSA research. This dataset can be used as a benchmark and facilitates further studies related to Sinhala natural language processing, and low-resource sentiment analysis tasks.
% Add some stats here
\end{abstract}


\section{Introduction}
Sentiment Analysis is a fundamental activity of Natural Language Processing (NLP) which involves the detection and understanding of subjective data in written format, such as opinions, emotions, and attitudes~\cite{pontiki2014semeval, pontiki2015semeval, pontiki2016semeval,gunathilaka2022aspect}. It has become a critical part of a broad spectrum of real-life applications such as product review analysis, social media monitoring, customer feedback analysis, and public opinion mining. Conventional methods of sentiment analysis often provide one sentiment label, e.g. positive, negative, or neutral, to a whole document or sentence~\cite{matlatipov2024uzabsa}. Although useful in a wide range of cases, this coarse-grained analysis may dramatically fail when it comes to handling complex opinions that are presented in the natural language given that there are many different things that can be said about an entity, and one can have varying degrees of sentiment~\cite{pontiki2014semeval}.

\begin{figure*}[!htb]
    \centering
    \includegraphics[width=0.95\textwidth]{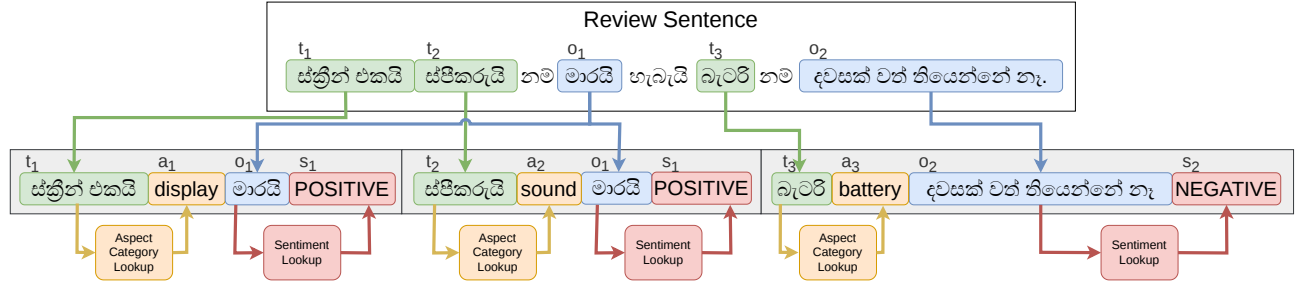}
    \caption{An example of three \texttt{(t, a, o, s)} sentiment quadruples being extracted from a single review sentence. Note how due to inflection in Sinhala language and hypernymy present in all languages, an assortment of \texttt{t} may get mapped to the same \texttt{a}.}
    \label{fig:example}
\end{figure*}

Aspect-based Sentiment Analysis (ABSA) has been suggested to overcome these shortcomings as a fine-grained sentiment analysis. Instead of creating a single overall sentiment label, ABSA aims to find specific aspects or features referred to in the text and the sentiment polarity that they have~\cite{pontiki2014semeval,gunathilaka2022aspect}. 
%ABSA does not create a single overall sentiment label, but aims to find specific aspects or features referred to in the text and the sentiment polarity that they have~\cite{pontiki2014semeval}.  
For example, a Sinhala product review may simultaneously express a positive opinion about a \texttt{product's display}  and \texttt{sound quality} while conveying a negative sentiment about its \texttt{battery life}, each representing a distinct aspect–sentiment pair, as illustrated in Fig \ref{fig:example}. ABSA allow breaking down opinions into aspect-sentiment pairs, which can be more detailed and actionable feedback on users. Consequently, the topic of ABSA has received high research activity, which has given rise to benchmark datasets and evaluation campaigns, especially on high-resource languages~\cite{pontiki2014semeval, pontiki2015semeval, pontiki2016semeval}.

Even though there is an abundance of ABSA datasets for high-resourced languages such as English and Chinese~\cite{rizvi2025enhancing}, there is no publicly available ABSA dataset for Sinhala that provides manually annotated aspect terms, opinion terms, and sentiment polarities at the quadruple level across e-commerce product domain~\cite{desilva2019survey}. The few publicly available sentiment analysis datasets for Sinhala such as that of~\citet{de2025geesanbhava}, have annotated the sentiments at the document level rather than the aspect level. 
%Even though activity of ABSA has already been researched in English and Chinese and some other languages rich in resources, there is a very limited amount of research on Sinhala Aspect-Based Sentiment Analysis~\cite{rizvi2025enhancing}. 

In order to fill this gap, we present, \dataSetName, the first publicly available Sinhala ABSA dataset with manual quadruple-level annotations comprising aspect terms, aspect categories, opinion terms, and sentiment polarities of standard Sinhala and code-mixed reviews in e-commerce product domain  \footnote{Hence the name \dataSetName, where \textit{Sal} means market or shops, \textit{Anga} means aspects, and \textit{Bhava} means sentiments in Sinhala}. The anotation were done according to a well-developed and linguistically informed annotation plan. Annotations were carried out by native Sinhala speakers with elaborate guidelines with the  aim of consistency and reliability. The dataset can be used to support several ABSA subtasks, such as aspect term extraction and aspect sentiment polarity classification, which allows pipeline-based and end-to-end modeling methods.

Beyond dataset building, we present an in-depth statistical analysis of the dataset. These benchmarks identify the inherent issues in Sinhalese ABSA and give reference points for future studies. Our goal is to support reproducible research by making the \dataSetName{} dataset is publicly available on \gh{https://github.com/lakshani-98/sinhala-absa-dataset}{GitHub} and \hf{https://huggingface.co/datasets/lakshani005/SalAngaBhava}{HuggingFace} to promote additional progress in the fine-grained sentiment analysis of Sinhala and other low-resource languages.  %is publicly available at \url{https://github.com/anon-absa-data/sinhala-absa-dataset}.

\section{Task Descriptions}

The presented data set can be used in tasks that constitutes Aspect Based Sentiment Analysis for Sinhala. It allows the fine-grained labelling of the Sinhala reviews at the aspect level. Given an input sentence, the task is to identify the aspects that are mentioned in the sentence and the sentiment polarity that corresponds to the aspect~\cite{yan2021unified}. The task formulation is based on the generally recognized ABSA settings~\cite{pontiki2014semeval, pontiki2015semeval, pontiki2016semeval}, which allow being compared with current benchmarks and customized to the language specifics of Sinhala~\cite{desilva2019survey}.

\subsection{Aspect Term Extraction}

Aspect Term Extraction (ATE) aims to identify explicit \textit{aspect terms} mentioned in a sentence. An aspect term refers to a word or phrase that denotes an attribute or component of an entity about which an opinion is expressed~\cite{yan2021unified}.
Formally, given a sentence $S = \{w_1, w_2, \ldots, w_n\}$, the goal of ATE is to extract a set of spans $A = \{a_1, a_2, \ldots, a_k\}$, where each $a_i$ corresponds to a contiguous sequence of tokens representing an aspect term~\cite{yan2021unified}.
% \textbf{Example:}
% \begin{quote}
% Sentence: {\sinhala මෙම දුරකථනයේ බැටරිය ඉතා හොඳයි.} \\
% Aspect Term: {\sinhala බැටරිය}
% \end{quote}
Following the conventions set by~\citet{maroof2024aspect}, this dataset is annotated with both explicit and implicit aspect terms. Explicit aspects are those explicitly expressed in the sentence, while implicit aspects are those implied by the sentence, even though they are not explicitly mentioned.

\subsection{Opinion Term Extraction}

The goal of the task of Opinion Term Extraction (OE) is to extract words or phrases that express opinions, sentiments, or other subjective feelings about an aspect or entity. Opinion terms are usually used to represent the emotional expressions used by an author~\cite{yan2021unified}.
Formally, given a sentence $S = \{w_1, w_2, \ldots, w_n\}$, the goal of OE is to extract a set of spans $O = \{o_1, o_2, \ldots, o_m\}$, where each $o_i$ corresponds to a contiguous sequence of tokens representing an opinion term~\cite{yan2021unified}.
This activity involves the identification of the expressive components of sentiment that are commonly used in combination with identified aspect terms, to gain more insight into the opinion structure of the sentence~\cite{xiao2021bert4gcn}.

\subsection{Aspect Level Sentiment Classification}

Aspect Level Sentiment Classification (ALSC) is about identifying the sentiment polarity for each aspect term in the context of an opinion term. For each aspect term found in a sentence, a sentiment polarity label is given
based on the sentiment expressed about the aspect~\cite{yan2021unified}. The set of sentiment labels used in this study is ~\cite{wan2020target}: \texttt{\{Positive, Negative, Neutral\}}. The sentiment polarity is determined by considering the local context around the aspect term and meaning of the sentence .

% \textbf{Example:}

% \begin{quote}
% Sentence: \textit{මෙම දුරකථනයේ බැටරිය ඉතා හොඳයි, නමුත් මිල වැඩියි.} \\
% Aspect--Sentiment Pairs: (\textit{බැටරිය}, Positive), (\textit{මිල}, Negative)
% \end{quote}

\subsection{Joint Aspect Based Sentiment Analysis}

Apart from the subtasks described above, the data allows joint Aspect-based Sentiment Analysis,
where aspect terms and sentiment polarities can be jointly predicted in an end-to-end manner. This task setting is motivated by real world applications, and enables researchers to investigate unified modelling approaches~\cite{wan2020target}. For the joint task, the output for a sentence is a list of aspect–sentiment pairs:$\{(a_1, p_1), (a_2, p_2), \ldots, (a_k, p_k)\}$, where $a_i$ denotes an aspect term and $p_i$ denotes its associated sentiment polarity.

\section{Related Work}
\subsection{Sentiment Analysis for Sinhala}
Sinhala NLP remains severely resource-constrained, as documented comprehensively by de Silva~\cite{desilva2019survey}. Early sentiment analysis work for Sinhala operated at the document level, with Senevirathne et al.~\cite{senevirathne2020sentiment} presenting a publicly available dataset of 15,059 Sinhala news comments annotated with positive, negative, neutral, and conflict labels using deep learning models such as RNN, LSTM, and Bi-LSTM.  De Mel and de Silva~\cite{de2025geesanbhava} similarly introduced GeeSanBhava, a document-level sentiment dataset for Sinhala music video comments. However, these works do not provide aspect-level annotations and cannot be used for ABSA tasks.

More recent work has extended Sinhala sentiment analysis to multilingual and code-mixed settings. Rizvi et al.~\cite{rizvi2025keyword} introduced a dataset of banking reviews in Sinhala, English, and code-mixed text, along with a keyword extraction and aspect classification system for banking reviews in Sinhala, English, and code-mixed content. They has fine-finetuned XLM-RoBERTa and achieved 87.4\% accuracy on Sinhala and code-mixed keyword extraction. A related study~\cite{rizvi2025enhancing} by the same group developed a hybrid ABSA framework with explainability using SHAP and LIME, achieving an F1-score of 0.89 for English and 88.4\% accuracy for Sinhala and code-mixed content.However, the dataset is manually annotated at the comment level, assigning predefined aspect categories and sentiment labels to whole comments without identifying explicit aspect terms or opinion terms, and is not publicly available. In contrast, \dataSetName{} covers the e-commerce domain, provides full quadruple-level manual annotations including target terms, opinion terms, aspect categories, and sentiment polarities, and is publicly released.

\subsection{ABSA Datasets for Low-Resource Languages}
ABSA research has been predominantly driven by high-resource language datasets, notably the SemEval series by Pontiki et al.~\cite{pontiki2014semeval, pontiki2015semeval, pontiki2016semeval} for English, which introduced the widely adopted aspect–sentiment pair annotation format. Comparable datasets have since been developed for other low-resource languages, such as UzABSA for Uzbek~\cite{matlatipov2024uzabsa} which follows a similar annotation methodology. \dataSetName{} follows and extends this tradition for Sinhala, adopting the quadruple annotation scheme proposed by Zhang et al~\cite{zhang2021aspect} to capture richer sentiment structure.

\subsection{Transliteration for Sinhala}
A notable preprocessing challenge in Sinhala NLP is the prevalence of Romanized Sinhala, where users write Sinhala in English script. De Mel and de Silva~\cite{de2025linguistic} have studied this phenomenon in the context of Sinhala YouTube comments, employing the Google Transliterator API to convert Romanized Sinhala to Unicode Sinhala script, successfully transliterating 28,043 out of 30,633 Romanized comments.

\section{Methodology}
This section describes the creation of \dataSetName, Sinhala Aspect-based Sentiment Analysis dataset, including the data collection and the annotation procedures. The aim of the dataset is to offer high-quality, fine-grained sentiment labels that reflect the sentiment expressed in Sinhala text.

\subsection{Data Collection and Preprocessing}
\label{sec:data:col}

The raw text data was collected from publicly accessible online resources of Sinhala user-generated text such as Daraz\footURL{https://www.daraz.lk/} using web scraping methods, similar to the data collection approach used by Viththalani et al.~\cite{patel2025aspect}. To ensure ethical compliance, only publicly visible reviews were collected, and no user-identifying information was extracted. The data collection were conducted with careful consideration of Daraz's terms of service and user privacy expectations. The data went through a multi-stage cleaning and normalisation process to improve its quality and consistency for the later stages of annotation and modelling. First, the reviews were automatically labeled into one of the following groups: (1) Pure Sinhala; Sinhala reviews in Sinhala script (Unicode), (2) Pure English; English reviews in English script, (3) Sinhala reviews in English script; also known as Romanised or \textit{Singlish}, and (4) Code Mixed; a mixture of Sinhala and English words written in the respective scripts.
% , as is often the case in online communication
%, which is frequently used in online platforms
%\begin{itemize}
%    \item Pure Sinhala (using Sinhala Unicode).
%    \item Pure English.
%    \item Sinhala using English Roman characters.
%    \item Code Mixed.
%\end{itemize}
This step was achived using a combination of Unicode recognition and English vocabulary matching. The reviews written in pure English were excluded from the dataset, as the aim of this study was to provide a resource for Sinhala. Pure Sinhala and code mixed reviews were kept as they where while the Romanised Sinhala reviews were transliterated in to Sinhala using the Google Transliterator API~\cite{de2025linguistic}. From the 10,989 cleaned reviews, 5,732 (51.9\%) were originally written in Romanized Sinhala and underwent transliteration via the Google Transliterator API prior to annotation. A further pre-processing step was performed to eliminate the non-textual items (such as special characters), text formatting and remove invalid (empty) entries. Any unknown and unidentified language patterns were also eliminated. Short reviews of one or two words were retained in the dataset as they represent a natural and common form of user expression in Sinhala e-commerce contexts. Despite their brevity, such reviews carry unambiguous sentiment polarity and can be meaningfully annotated at the aspect level with an implicit overall satisfaction aspect, contributing valid training signal for sentiment classification tasks.
Following the convention set by~\citet{pontiki2014semeval}, the resultant dataset is presented as sentence instances, which are processed as an individual unit for annotation. This design choice 
%aligns with established Aspect Based Sentiment Analysis benchmarks and 
enables support for both aspect term extraction and aspect sentiment polarity classification tasks.

\subsection{Annotation Process}

% https://app.diagrams.net/#G1k4P2nX-AXz22FnF5sXIUBWKME0Gh1uBn#%7B%22pageId%22%3A%22YoBKFaFIHoHEw3nqQ0CH%22%7D

The annotation process was carried out manually to ensure high annotation quality and linguistic accuracy. All annotations were performed by native Sinhala speakers with proficiency in language understanding and familiarity with sentiment analysis concepts. Prior to the annotation phase, detailed annotation guidelines were developed to clearly define aspect terms, sentiment polarity labels, and edge cases following a similar approach by Matlatipov et al~\cite{matlatipov2024uzabsa}.
As shown in Fig~\ref{fig:example}, we opted to use the \texttt{(t, a, o, s)} sentiment quadruple proposed by~\citet{zhang2021aspect} where:  
\begin{itemize}
    \item \textbf{\texttt{t} (Target Term):} The word or phrase representing the aspect, as it appears in the text. If the aspect is implicit, this is set to \texttt{NULL}.
    \item \textbf{\texttt{a} (Aspect):} The aspect category chosen from a list of aspects that are pre-defined in the dataset following the aspect inventory system proposed by~\citet{pontiki2016semeval}.
    \item \textbf{\texttt{o} (Opinion Term):} The smallest sentiment-bearing element of the aspect, expressing the opinion for a particular aspect in the sentence, as it appears in the text.
    \item \textbf{\texttt{s} (Sentiment):} The sentiment polarity of the aspect (one of the values: \texttt{positive, negative, neutral}) according to the opinion.
\end{itemize}
Note that the traditionally annotated~\cite{pontiki2014semeval, pontiki2015semeval, pontiki2016semeval} \textit{Aspect Term} is instead split into \textit{Target Term} and \textit{Aspect} in this scheme. The reason for this is the fact that Sinhala is a highly inflicted language~\cite{de2025linguistic}, thus the same \textit{Aspect} may have numerous lexical forms which, if used as a singular tag, will reduce the accuracy of downstream tasks by the numeracy of labels (reduces support per label) as well as the possibility of the labeled training set not being exhaustive. Further note that, for the rest of this paper, when we mention \textit{Aspect Term}, we mean the combined tuple \texttt{(t, a)}, which when taken together, is analogous to the classical definition of an \textit{Aspect Term}. 
%Each sentence was annotated in two stages. First, target terms mentioned in the sentence were identified and marked. Target terms were defined as words or phrases that denote specific attributes or components of an entity toward which an opinion is expressed~\cite{pontiki2014semeval}. In the second stage, a sentiment polarity label: \textit{Positive}, \textit{Negative}, or \textit{Neutral}  was assigned to each identified aspect term based on the contextual sentiment expressed in the sentence~\cite{pontiki2014semeval}.
%The sentences were manually labeled in the form of a quadruple (t, a, o, s)~\cite{zhang2021aspect} where:

To ensure annotation consistency, an initial pilot annotation was conducted on a subset of the data, followed by discussions among annotators to resolve ambiguities and refine the guidelines, in line with the methodology described by Hua et al.~\cite{hua2025edurabsa}. Disagreements during the main annotation phase were resolved through discussion and consensus, ensuring that the final annotations reflect a consistent interpretation of aspect-level sentiment.

This annotation strategy results in a high-quality dataset suitable for training and evaluating both pipeline-based and end-to-end Aspect Based Sentiment Analysis models for Sinhala. To assess annotation quality, an inter-annotator agreement (IAA) study was conducted on a randomly sampled subset of 100 reviews, independently annotated by 3 native Sinhala speakers following the same guidelines. Following the approach of Saeidi et al.~\cite{saeidi2016sentihood}, Cohen's Kappa~\cite{cohen1960coefficient} was computed over aspect-sentiment pairs across all three annotator pairs. The average pairwise Cohen's Kappa at the quad level was 0.82, indicating a good agreement. Aspect category agreement, measured using Jaccard similarity~\cite{jaccard1912distribution}, averaged 0.63, reflecting the inherent variability in aspect identification in fine-grained ABSA tasks. This is expected given the diversity of aspect categories and the implicit nature of many Sinhala e-commerce reviews. These results confirm the reliability and consistency of the SalAngaBhava annotations.

\subsection{Scope and Assumptions}

We have followed these assumptions listed below when annotating the dataset to ensure consistency in annotation, while capturing a broader range of aspect expressions, and facilitate model evaluation across a variety of modeling techniques:
\begin{itemize}
    \item Aspect terms are annotated whether they are explicitly or implicitly mentioned. %Explicit aspects are explicitly mentioned in the sentence, while implicit aspects are implied.
    \item Each aspect term is given a single sentiment polarity (positive, negative or neutral).
    \item When there are conflicting sentiments expressed for the same aspect in a sentence, the sentiment is determined by the most dominant sentiment expressed.
    \item Aspect annotations are more focused on semantics rather than word forms so that paraphrases and slang expressions are mapped to the same aspect.
    \item Neutral is given when the opinion is neither positive nor negative.
\end{itemize}

%These design decisions ensure consistency in annotation, while capturing a broader range of aspect expressions, and facilitate model evaluation across a variety of modeling techniques.

\subsection{Structure of the Dataset}
The \dataSetName{} dataset is released in a machine-readable and researcher-friendly format to facilitate reproducibility and reuse in future studies.
%\subsubsection{Dataset Structure}
Each instance in the dataset contains the following fields:

\begin{itemize}
     \item \textbf{sentence\_id}: A unique identifier assigned to each sentence for reference and traceability.
    \item \textbf{product\_name}: The name of the product being reviewed.
    \item \textbf{rating}: The numerical rating assigned by the user to the product (e.g., on a 1--5 scale).
    \item \textbf{price}: The listed price of the product at the time of review, when available.
    \item \textbf{review}: The original review text written in Sinhala, which may include native Sinhala script or Romanized Sinhala.
    \item \textbf{cleaned\_review}: The processed version of the review after cleaning and transliteration, where applicable.
    \item \textbf{review\_type}: The classification of the review indicating whether it is \textit{pure\_sinhala}, \textit{sinhala\_in\_english} or \textit{code\_mixed}.
    \item \textbf{aspect\_sentiment\_annotation}: For each annotation, a \texttt{(t, a, o, s)} sentiment quadruple as proposed by~\citet{zhang2021aspect}. 
    %A unified annotation field containing both the aspect category and its corresponding sentiment polarity. Each annotation is represented in the structured form $(t, a, o, s)$ as defined in the annotation schema, where the aspect and sentiment are jointly encoded rather than stored in separate columns~\cite{zhang2021aspect}.
     \item \textbf{category}: The category of the product.
\end{itemize}

%\subsubsection{Encoding and Language Representation}
All textual content is encoded using UTF-8 Unicode encoding. Sinhala text is represented using native Sinhala characters to preserve linguistic integrity. 
As discussed in Section~\ref{sec:data:col}, reviews originally written in English letters (Romanized Sinhala), a transliteration step is applied to convert them into standard Sinhala Unicode script. The final dataset only contains Unicode Sinhala text to ensure consistency.

The originally collected dataset consisted of 17,500 reviews, which was reduced to 10,989 after preprocessing and cleaning. This means, \dataSetName, which we create as the first public dataset for Sinhala ABSA is larger than that of \citet{pontiki2014semeval} which was a foundational dataset for English ABSA.
A subset of 1858 reviews from the cleaned dataset have been human-annotated at the aspect-level sentiment. The full dataset is publicly released, comprising both the annotated subset (1,858 reviews) and the remaining unannotated reviews, totalling 10,989 cleaned reviews.
%, while annotation is still in progress for all product categories. 
The number of  reviews is imbalanced across categories, with \textit{Electronic Products}, \textit{Skincare} and \textit{Home Appliances} dominating, while the categories of \textit{Fashion} and \textit{Grocery} have fewer reviews. This distribution is representative of the distribution of user reviews on e-commerce platforms. The distribution of reviews and annotation coverage in each category is shown in Table~\ref{tab:dataset_distribution}. The annotation coverage also varies considerably across categories, ranging from 4.28\% of Electronic Products reviews to 100\% of Fashion reviews. Within each category, reviews were randomly sampled for annotation, with the number of annotated reviews per category determined by available annotator hours. Smaller categories such as Fashion were fully annotated due to their limited size, while larger categories such as Electronic Products could only be partially covered due to the resource and time constraints. Although this introduces uneven coverage across categories, the random sampling within each category ensures that the annotated subset is a representative sample of the available reviews, mitigating category-level sampling bias. We acknowledge this as a limitation to be addressed in future extensions of the dataset.

\begin{table}[h]
\centering
\caption{Distribution of reviews across product categories and annotation coverage}
\label{tab:dataset_distribution}
%\begin{tabular}{lrrr}
\begin{tabularx}{0.48\textwidth}{lRRR}
\hline
\multirow{2}{*}{\textbf{Category}} & \multicolumn{2}{c}{\textbf{Reviews}} & \textbf{Coverage} \\
\hhline{~--~}
& \textbf{Total} & \textbf{Annotated} & \textbf{(\%)}\\
\hline
Electronic Products & 5961 & 255 & 4.28\\
Home Appliances & 1833 & 700 & 38.18\\
Grocery & 986 & 400 & 40.57\\
Fashion (Clothing and Shoes) & 153 & 153 & 100 \\
Skincare & 2056 & 350 & 17.02\\
\hline
\textbf{Total} &  \textbf{10989} & \textbf{1858} & \textbf{16.91}\\
\hline
%\end{tabular}
\end{tabularx}
\end{table}

\section{Evaluation Measures and Baselines}

This section outlines the baseline experiments that were run using statistical and embeddings-based approaches to understand dataset characteristics and ensure its quality.
%\subsection{Evaluation of the Dataset}
%This section presents a comprehensive analysis of the proposed dataset to assess its quality, diversity and representativeness. 
The analyses focus on aspect distribution, sentiment balance, and semantic consistency to ensure the dataset is suitable for robust Aspect Based Sentiment Analysis tasks.

\subsection{Aspect Coverage and Distribution Analysis}

Following~\citet{lee2026dimabsa}, we analyzed the aspect category distribution  of our dataset. Aspect Coverage and Distribution Analysis involve assessing the distribution of aspect categories in the dataset and whether it covers a broad spectrum of opinion targets in the targeted domain. It helps determine whether the dataset includes a rich variety of aspect categories or is skewed towards a few common aspects. It also reveals long-tail aspects, which are infrequent but may represent more specific or domain-specific aspects. This analysis is vital for evaluating the realism of a dataset and ensuring models trained on the data have exposure to both common and rare aspects.

\begin{figure*}[!htb]
    \centering
    \includegraphics[width=\linewidth]{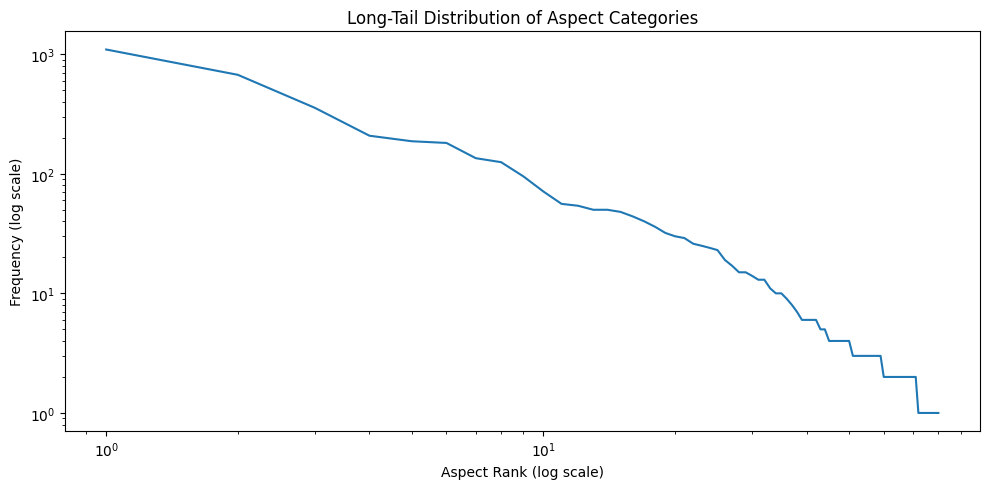}
    \caption{Long-Tail Distribution of Aspect Categories}
    \label{fig:LongTail}
\end{figure*}

\begin{table}[!htb]
\centering
\caption{Top 10 Aspect Distribution and Coverage Analysis}
\label{table:top_aspects}
\begin{tabular}{lrr}
\hline
\textbf{Aspect} & \textbf{Frequency (N)} & \textbf{Percentage (\%)} \\ \hline
Overall Satisfaction & 1,094 & 27.56\% \\
Product Quality & 672 & 16.93\% \\
Price / Value for Money & 356 & 8.97\% \\
Packaging Quality & 208 & 5.24\% \\
Delivery \& Packaging & 187 & 4.71\% \\
Delivery Speed & 181 & 4.56\% \\
Functionality & 135 & 3.40\% \\
Delivery Condition & 125 & 3.15\% \\
Performance & 95 & 2.39\% \\
Authenticity & 71 & 1.79\% \\ 
\hline
\end{tabular}
\end{table}       

\begin{figure*}[!htb]
    \centering
    \includegraphics[width=1\linewidth]{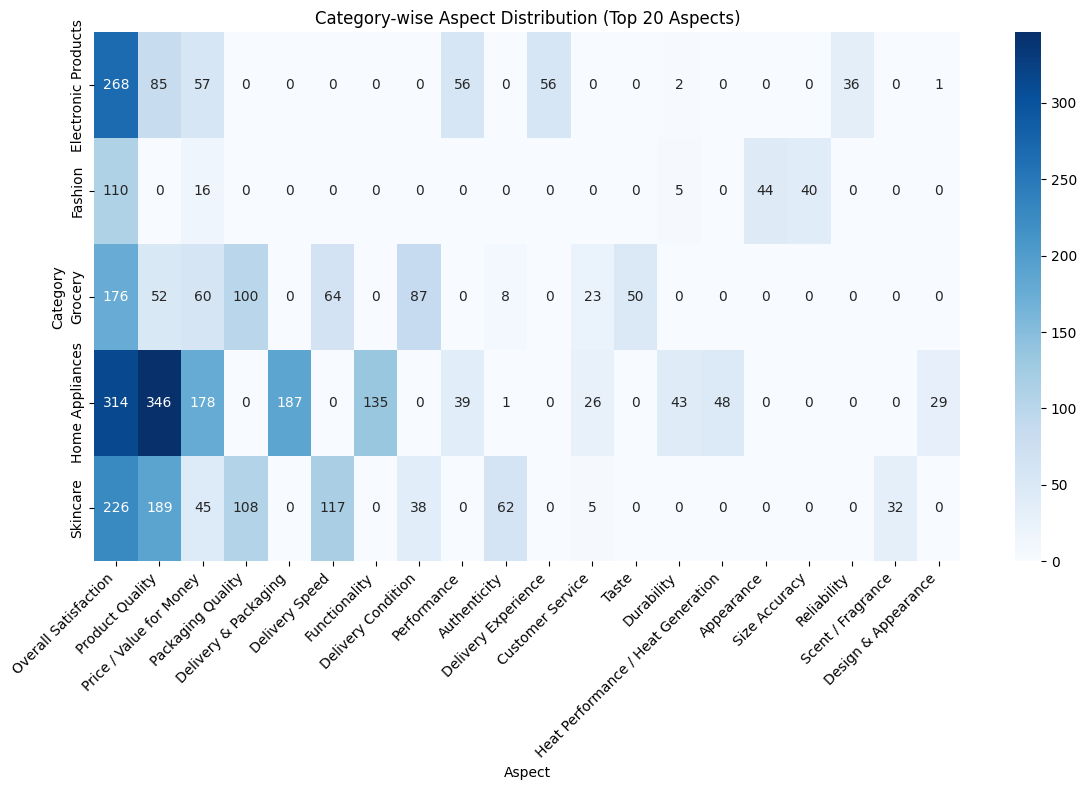}
    \caption{Category-wise Aspect Distribution (Top 20 Aspects)}
    \label{fig:CategoryWise}
\end{figure*}

The \dataSetName{} dataset has a highly skewed but a realistic long tail as demonstrated in Fig~\ref{fig:LongTail}, where some general evaluative features such as \textit{Overall Satisfaction}, \textit{Product Quality}, and \textit{Price} are highly over-represented. As shown in Table~\ref{table:top_aspects}, these three general aspects constitute 53.46\% of all annotations, which reflects the preference of consumers for essential product attributes. This trend is also illustrated in the Fig~\ref{fig:CategoryWise}, which demonstrates that while general aspects are common across all categories, fine-grained features are category specific, such as \textit{Scent/Fragrance} in \textit{Skincare} and \textit{Heat Performance} in \textit{Home Appliances}. This distribution guarantees that the dataset offers both a substantial amount of data for identifying common aspects and a \textit{tail} of infrequent features specific to a domain, necessary for the fine-grained ABSA models.

\subsection{Sentiment Polarity Balance Analysis}

Following~\citet{pontiki2014semeval}, we analyzed sentiment polarity balance of the dataset. This analysis examines the balance of sentiment polarities for aspect terms in the dataset, where it checks if all sentiment classes are present sufficiently or if the dataset is skewed towards a specific sentiment class. Review data typically tends to be biased towards positive sentiments in the real world. The analysis of sentiment distribution is crucial as it can impact the performance of models, especially those evaluated with metrics like macro-averaged F1-score, which account for class imbalance. An evenly distributed or realistically skewed sentiment labels in the dataset is important to ensure that models built on the dataset reflect the true distribution of sentiment in the real world, and that fair evaluation is possible.

\begin{figure*}[!htb]
    \centering
    \includegraphics[width=1\linewidth]{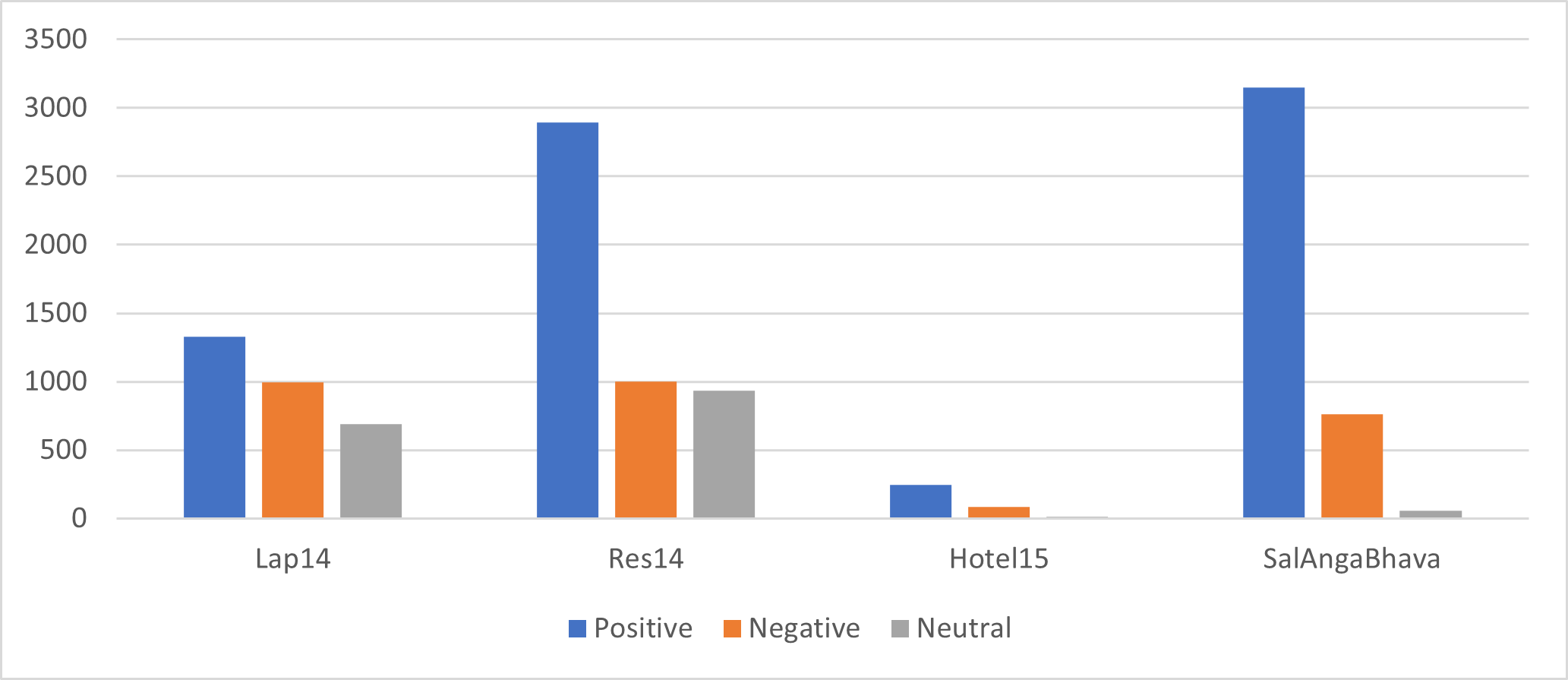}
    \caption{Sentiment Polarity Distribution of our \dataSetName{} compared to Res14~\cite{pontiki2014semeval}, Lap14~\cite{pontiki2014semeval} and Hotel15~\cite{pontiki2015semeval}}
    \label{fig:Sentiment}
\end{figure*}

\begin{figure}[!htb]
    \centering
    \includegraphics[width=1\linewidth]{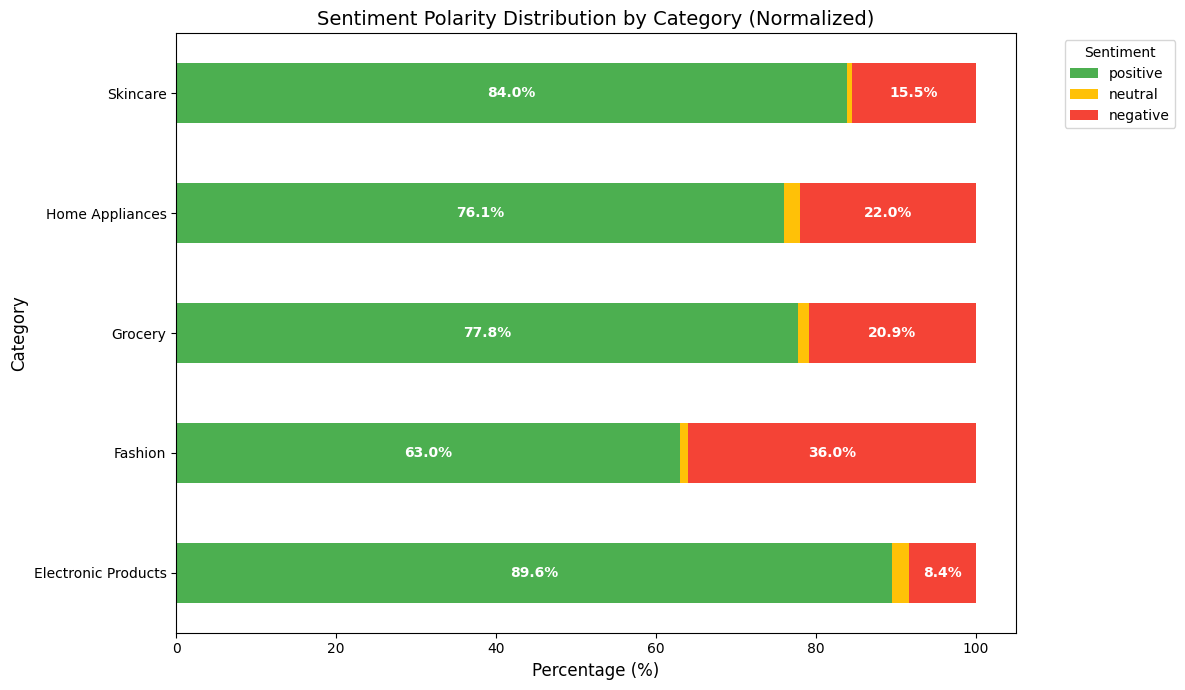}
    \caption{Sentiment Polarity Distribution by product Category (Normalized)}
    \label{fig:SentimentBy}
\end{figure}

The overall sentiment distribution is realistic. It reflects the class imbalance normally confronted in e-commerce data. It consists of 
3150 positive labels, 762 and 58 negative and neutral labels, respectively. As Fig~\ref{fig:Sentiment} shows, this general bias is complemented by the bias specific to category in Fig~\ref{fig:SentimentBy} as well. Although \textit{Electronic Products} are heavily skewed towards the positive side (89.6\%), \textit{Fashion} offers a more diverse view with 36.0\% negative sentiment. This variability is vital for ABSA model assessment, as it demands that they adapt to different degrees of class imbalance and be able to detect user dissatisfaction across a range of product domains. 

The neutral class is notably underrepresented in the dataset, with only 58 neutral labels compared to 3,150 positive and 762 negative labels.  This reflects the natural distribution of user-generated e-commerce reviews, where users tend to express clear positive or negative opinions rather than neutral reviews. The neutral class remains valid as a sentiment class for all ABSA tasks, but its low frequency could be problematic for other models aiming to classify neutral sentiment. To overcome this, we suggest the macro-averaged F1-score as the main  evaluation metric for downstream tasks on \dataSetName{} since it is not influenced by class imbalance and gives a more useful score of model performance in the presence of class imbalance. Collecting additional neutral annotations remains a priority for future extensions of the dataset.

\subsection{Pairwise Similarity Comparison} 
To explore the semantic coherence and structural quality of the proposed dataset, we conducted a Pairwise Similarity Comparison using the Language-Agnostic BERT Sentence Embeddings (LaBSE)~\cite{feng2022language} vector representations of the dataset, following the method used by~\citet{peiris2022synthesis}. The data was represented using the aspect - opinion pairs. And we computed the cosine similarity between randomly selected pairs under three scenarios: (i) same aspect - same sentiment, (ii) same aspect - different sentiment, and (iii) different aspects.

\begin{figure}[!htb]
    \centering
    \includegraphics[width=\linewidth]{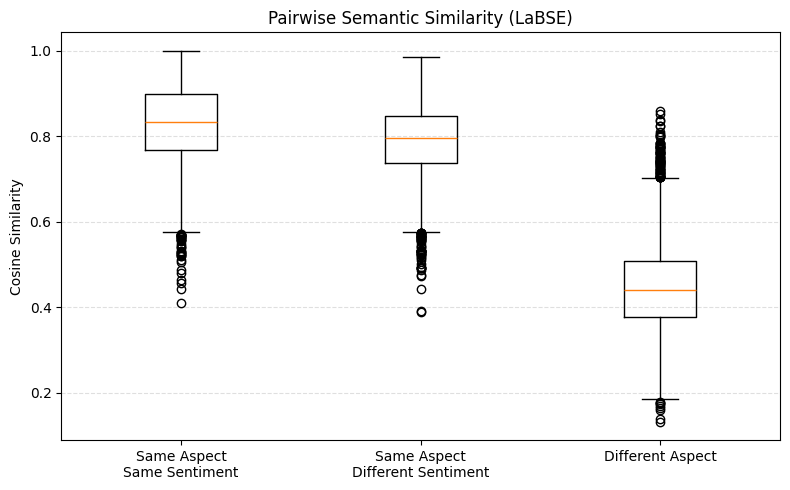}
    \caption{Pairwise Semantic Similarity (LaBSE)}
    \label{fig:Pairwise}
\end{figure}

The pairwise similarity analysis based on vector embeddings from LaBSE attests the semantic coherence as well as the consistency of this dataset. As the box plot in Fig~\ref{fig:Pairwise} shows, the pairs within the same aspect category have a high median cosine similarity, irrespective of sharing the same or opposite sentiment. It is average around 0.83 for same and 0.80 for opposite. This proves that the vector representations competently represent the semantic similarities between aspects. But then, the median similarity between different aspects is much lower (around 0.44) compared to above two scenarios. This distinct intra-aspect and inter-aspect correlations grouping confirms the ability of the model to distinguish between different categories as well as the semantic distinctiveness of the dataset. This establishes that the dataset can be used as an excellent resource to train highly accurate ABSA models as well.

\subsection{Product Category Classification}
We use product category classification to assess the quality and distinctiveness of the existing category labels. If a model trained on the reviews can accurately predict the category labels, then this provides confidence that the categories are linguistically separable and consistently labelled. Following~\citet{oancea2023text}, we build a TF-IDF with Logistic Regression baseline, and a FastText~\cite{bojanowski2017enriching} model, using the existing product category labels as ground truth, to investigate the extent to which the textual reviews match their product category labels.

\begin{table}[!htb]
\centering
\caption{Product Category Classification Performance Comparison}
\label{table:classification_results}
\begin{tabular}{l>{\centering\arraybackslash}p{1.5cm}>{\centering\arraybackslash}p{1.5cm}r}
\hline
\textbf{Category} & \textbf{TF-IDF F1-Score} & \textbf{FastText F1-Score} & \textbf{Support} \\ \hline
Electronics & 0.74 & 0.77 & 1,817 \\ 
Fashion & 0.31 & 0.35 & 73 \\ 
Grocery & 0.45 & 0.54 & 471 \\ 
Home Appliances & 0.25 & 0.43 & 651 \\ 
Skincare & 0.29 & 0.42 & 413 \\ 
\hline
\textbf{Overall Accuracy} & \textbf{0.60} & \textbf{0.64} & \textbf{3,425} \\ 
\textbf{Macro Avg F1} & \textbf{0.41} & \textbf{0.50} & \textbf{3,425} \\\hline 
\end{tabular}
\end{table}

%\section{Evaluation Results}
%This section presents an analysis of the dataset across multiple evaluation dimensions, providing insights into aspect coverage, domain diversity, and sentiment distribution. These analyses help assess the overall quality and suitability of the dataset for Aspect Based Sentiment Analysis tasks.
%\subsection{Aspect Coverage and Distribution Analysis}
%\subsection{Sentiment Polarity Balance Analysis}
%\subsection{Pairwise Similarity Comparison}
%\subsubsection{Product Category Classification}

%We benchmarked the classification of the categories, using a TF-IDF with Logistic Regression baseline model and a FastText model to assess the distinction and separability of the categories. 
The findings suggest that the reviews has the linguistic distinctiveness to enable automatic classification, with FastText performing better than the IF-IDF with Logistic Regression, with a higher overall accuracy of 63.85\% versus 60.32\% as presented in the Table~\ref{table:classification_results}.

Further analysis of the class-specific metrics shows that \textit{Electronic Products} is the most distinctive category, with an F1-score of 0.77 in the FastText model, potentially due to its specialised technical terms. In contrast, \textit{Home Appliances} and \textit{Skincare} have lower recall and F1-scores, likely due to some overlapping or \textit{vocabulary leakage} between these classes.

Additionally, the substantial improvement in the macro-averaged F1-score of FastText (0.50 vs. 0.41) suggests that the sub-word information and embedding representation is more effective at capturing the distinguishing features of the smaller classes than the sparse TF-IDF feature vector. This gives us confidence that while the category labels are linguistically distinct, the dataset is a challenging classification task, and accurately reflects the overlap of different e-commerce product descriptions.

\section{Baseline Results}
To validate \dataSetName{} for ABSA tasks, we report baseline results for Aspect Term Extraction (ATE) and Aspect Level Sentiment Classification (ALSC) using InstructABSA~\cite{scaria2024instructabsa}, an instruction-tuned model based on mT5-small. Table~\ref{table:baseline_results} reports the results.
\begin{table}[h]
\caption{Baseline Results — InstructABSA, Macro F1}
\label{table:baseline_results}
\centering
\begin{tabular}{p{2cm}p{2cm}p{2cm}}
\hline
 & \textbf{ATE} & \textbf{ALSC} \\
\hline
\textbf{Macro F1} & 0.72 & 0.28 \\
\hline
\end{tabular}
\end{table}

\section{Conclusion}

We present the \dataSetName{}  dataset, a novel, language driven resource to support fine-grained sentiment analysis in Sinhala language. By creating a dataset of annotated aspect terms and their associated sentiment polarities in 5 different e-commerce categories, this paper contributes to the low-resource NLP which is an under-researched field. The dataset was constructed with ethical considerations in mind. Only publicly visible reviews in Daraz were used, and no personally identifiable information was collected. Inter-annotator agreement experiments on a sample of 100 reviews confirmed the reliability of the annotations, with an average Cohen's Kappa of 0.82 at the quadruple level.

The further evaluation experiments confrim that the dataset is well structured and reflects the real world complexities of the consumer behavior.The similarity analysis shows a consistent semantic structure as well as classification task confirms the distinctiveness. These results together confirm that the dataset offers distinct, separable cues for machine learning models. It also maintains the diversity of natural language. 

The distribution of the data shows a well-balanced blend of common evaluative features with the difficult long tail of domain-specific attributes. That makes the dataset suitable for a wide range of ABSA tasks. Finally, the diverse sentiment polarities across multiple different product domains offer a strong basis for evaluating model resilience to class imbalance. In conclusion, this dataset offers a solid foundation for building advanced sentiment models, contributing to the advancement of NLP for Sinhala and the resource gap for low-resource languages in the global NLP ecosystem.

{\footnotesize
\bibliographystyle{IEEEtranN}
\bibliography{references}
}
\end{document}